\documentclass[10pt,journal,compsoc]{IEEEtran}

\ifCLASSOPTIONcompsoc
  \usepackage[nocompress]{cite}
\else
  \usepackage{cite}
\fi

\ifCLASSINFOpdf
  \usepackage[pdftex]{graphicx}
  \graphicspath{{../pdf/}{../jpeg/}}
  \DeclareGraphicsExtensions{.pdf,.jpeg,.png}
\else
\fi

\usepackage{amsmath}
\usepackage{multirow}
\usepackage{array}
\usepackage{url}
\usepackage{hyperref}
\usepackage{booktabs}
\begin{document}
\title{NameRec\textsuperscript{*}: Highly Accurate and Fine-grained Person Name Recognition}
%
%
%
%

\author{Rui~Zhang*,
        Yimeng~Dai,
        and~Shijie~Liu
\IEEEcompsocitemizethanks{
    \IEEEcompsocthanksitem Rui Zhang~~* Corresponding Author\protect\hfil\break
    URL \url{https://ruizhang.info/} \protect\hfil\break
     rui.zhang@ieee.org
    \IEEEcompsocthanksitem Yimeng Dai\protect\hfil\break
    The University of Melbourne\protect\hfil\break
        yimengd@student.unimelb.edu.au
          \IEEEcompsocthanksitem Shijie Liu \protect\hfil\break
          The University of Melbourne\protect\hfil\break
          shijiel2@student.unimelb.edu.au
    }
}



\markboth{Zhang \MakeLowercase{\textit{et al.}}: NameRec\textsuperscript{*}: Highly Accurate and Fine-grained Person Name Recognition}
{Zhang \MakeLowercase{\textit{et al.}}: NameRec\textsuperscript{*}: Highly Accurate and Fine-grained Person Name Recognition}

%


\IEEEtitleabstractindextext{%
\begin{abstract}
  In this paper, we introduce the NameRec* task, which aims to do highly accurate and fine-grained person name recognition. Traditional Named Entity Recognition models have good performance in recognising well-formed person names from text with consistent and complete syntax, such as news articles. However, there are rapidly growing scenarios where sentences are of incomplete syntax and names are in various forms such as user-generated contents and academic homepages. To address person name recognition in this context, we propose a fine-grained annotation scheme based on anthroponymy. To take full advantage of the fine-grained annotations, we propose a Co-guided Neural Network (CogNN) for person name recognition. CogNN fully explores the intra-sentence context and rich training signals of name forms. To better utilize the inter-sentence context and implicit relations, which are extremely essential for recognizing person names in long documents, we further propose an Inter-sentence BERT Model (IsBERT). IsBERT has an overlapped input processor, and an inter-sentence encoder with bidirectional overlapped contextual embedding learning and multi-hop inference mechanisms. To derive benefit from different documents with a diverse abundance of context, we propose an advanced Adaptive Inter-sentence BERT Model (Ada-IsBERT) to dynamically adjust the inter-sentence overlapping ratio to different documents. We conduct extensive experiments to demonstrate the superiority of the proposed methods on both academic homepages and news articles.

\end{abstract}

}

\maketitle

\IEEEdisplaynontitleabstractindextext

%
\IEEEpeerreviewmaketitle

\IEEEraisesectionheading{\section{Introduction}\label{sec:introduction}}
Named Entity Recognition aims to locate and classify named entities from unstructured text into predefined categories such as person names, locations, organizations, quantities, time, medical codes, etc. Among the named entities, person names are one of the most important and person name recognition serves as the basis for many downstream applications in knowledge management and information management.
Recognising person names from unstructured text is an important process for many online academic mining system, such as AMiner \cite{tang2008arnetminer} and CiteSeerX \cite{ororbia2015big}. The recognized person names plays an important role for knowledge base enrichment \cite{trisedya2019neural}.
Person name recognition can also provide valuable insights for learning the relationships between people and provides valuable insights for analysing their collaboration networks. \cite{barrio2014reel,liu2014full}.

\begin{figure}[!th]
\includegraphics[width=0.94\columnwidth]{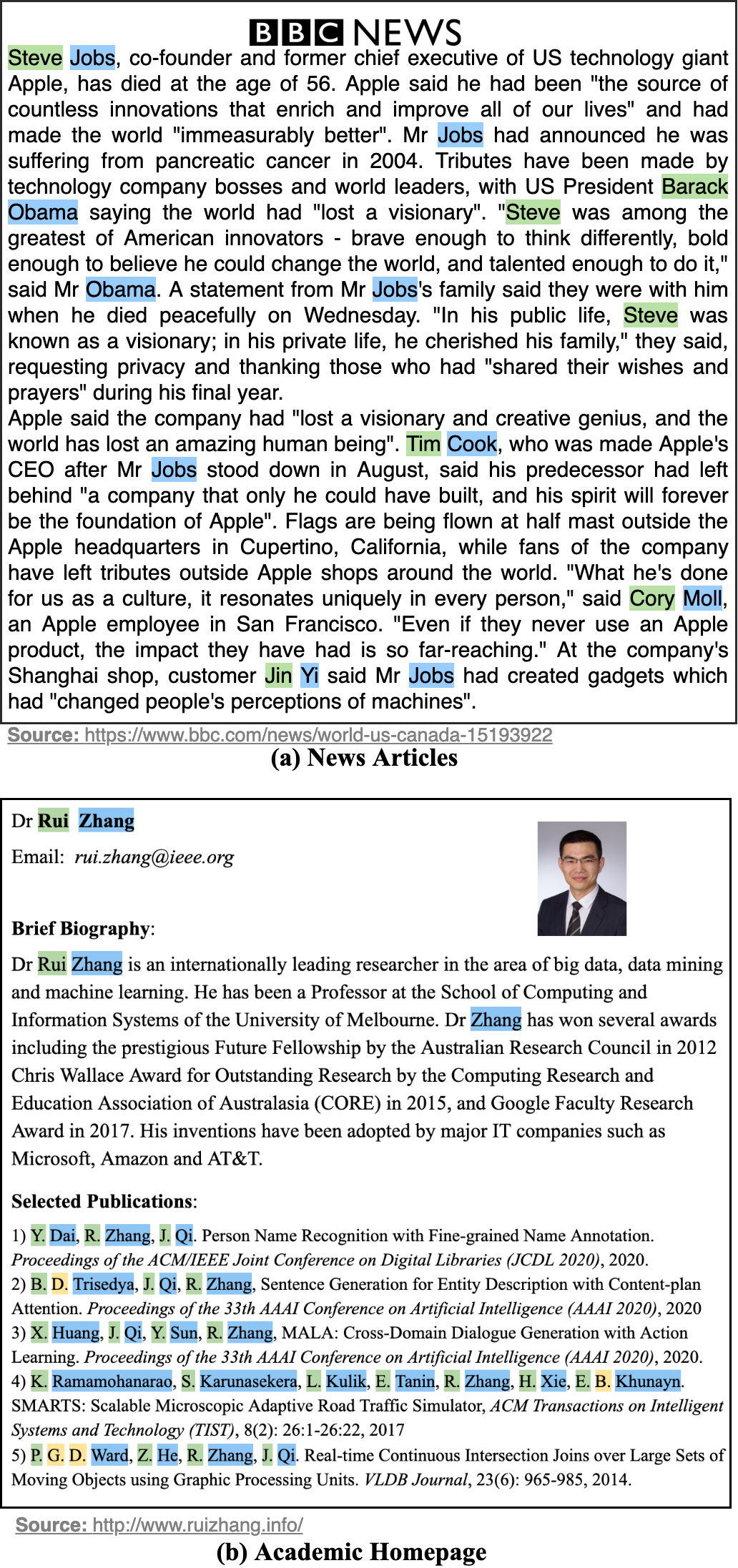}
\caption{An example of a news article with well-formed text and an example of an academic resume with free-form text. All the person names are highlighted.} \label{news_cv}
\end{figure}

Traditional NER models \cite{huang2015conll,chiu2016conll,ma2016conll} have achieve good performance on recognising person names from well-formed text with consistent and complete syntax, such as news articles (cf. Figure~\ref{news_cv}(a) ). Person names in such text often have straightforward patterns, e.g., first name in full followed by last name in full. However, challenges remain for recognising person names from free form text. These may appear in many applications, such as user-generated academic homepages, resumes, articles in online forums and social media (cf. Figure~\ref{news_cv}(b)). These text often contain person names of various forms with incomplete syntax.

\begin{figure*}[!t]
\includegraphics[width=\textwidth]{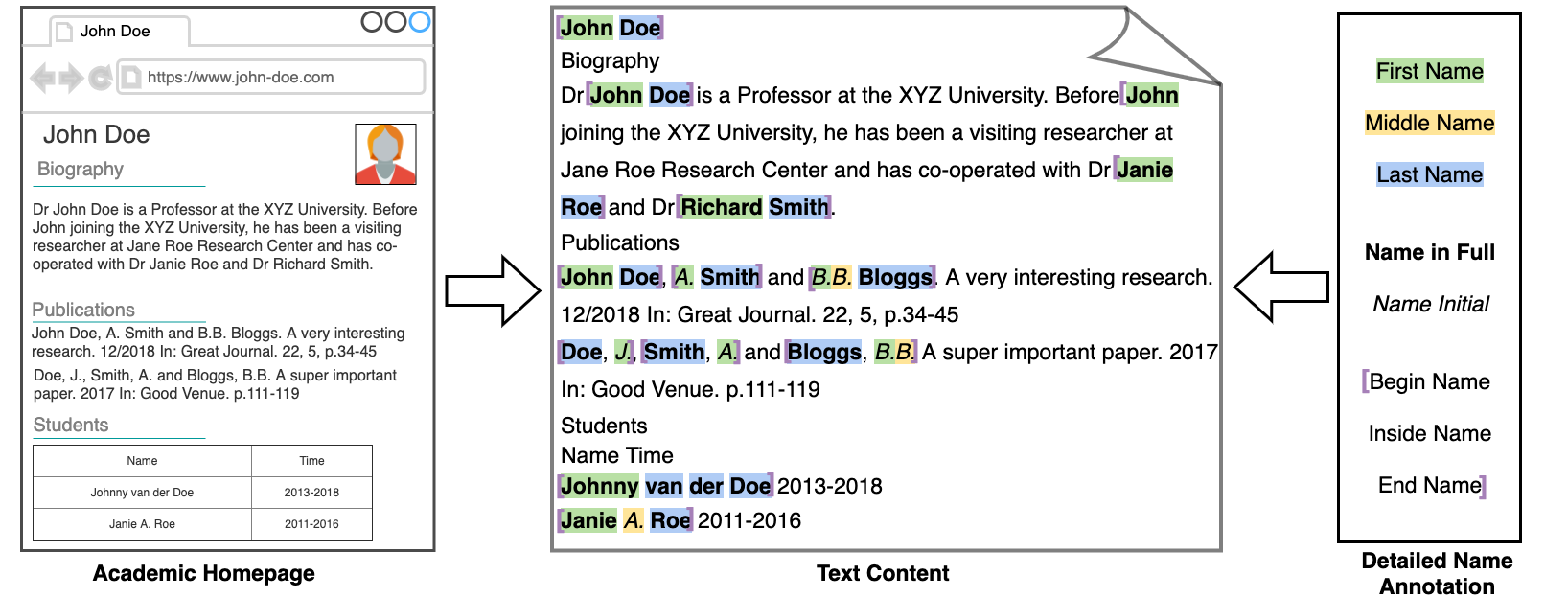}
\caption{An example of person name recognition in academic pages. All person names are highlighted (best viewed in color).} \label{web_example}
\end{figure*}

In this paper, we introduce the NameRec* task, which aims to do highly accurate and fine-grained person name recognition. Figure \ref{web_example} shows an example of person name recognition in academic homepages. The biography section consists of complete sentences while the students section simply lists information in a line. The person names may be in different forms. Figure \ref{web_example} contains well-formed full name of the researcher `\texttt{John Doe}' in the page header and abbreviated names in the publications section. Further, the abbreviated names may have different abbreviation forms, e.g., `\texttt{B.B. Bloggs}' vs. `\texttt{Doe, J.}'.

To better recognize person names from such a free-form text,
we exploit knowledge from \emph{anthroponymy} \cite{felecan2012name} (i.e., the study of the names of human beings) and propose a fine-grained annotation scheme that labels detailed name forms which distinguishes first, middle, or last name, and full names or initials (cf.~Figure~\ref{web_example}).  Such fine-grained annotations offer richer training signals to NER models to learn the patterns of person names in free-form text. However, fine-grained annotations also bring challenges because more label classes need to be learned.

To take full advantage of the fine-grained annotations, we propose a multi-task learning model called \emph{Co-guided Neural Network} (CogNN) for person name recognition. CogNN consists of two sub-neural networks, which are Bi-LSTM-CRF variants. One sub-network focuses on predicting whether a token is a name token, while the other focuses on predicting the name form of the token. The intuition is that knowing whether a token is a part of a name helps recognise the fine-grained name form class of the token, and vice versa. For example, if a token is not considered as a part of a name, then even if it is a word initial, it should not be labelled as a name initial. However, the underlying correlation between different annotations cannot be captured well when the two sub-networks are trained together by simply minimising the total loss. The reason is that the learning signals of the two sub-neural networks are not shared sufficiently. To better capture the underlying correlation between different annotations, we share the learning signals of two sub-neural networks through a co-attention layer. Further, we use a gated-fusion layer to balance the information learned from two sub-networks. This way, neither sub-network is overwhelmed by misleading signals that may be learned by the other sub-network.

CogNN fully explores the intra-sentence context and rich training signals of name forms. However, the inter-sentence context and implicit relations, which are essential for recognizing person names in long documents, are not captured. For example, in Fig~\ref{web_example}, the sentence `\texttt{Dr John Doe is a Professor at the XYZ University}' in the academic homepage contains rich contextual information. The intra-sentence context indicates that the words `\texttt{John Doe}' may be a person's name, since the words appear after a title `\texttt{Dr}' and before a job occupation `\texttt{a Professor}'. However, in later part of the academic homepage, `\texttt{John Doe}' appears with much less intra-sentence context and is harder to recognise.

To address this issue, we propose a \emph{Inter-sentence BERT Model} (IsBERT) to capture the inter-sentence context and implicit relations. A naive BERT model\cite{devlin2019bert} only learns context-dependent word embeddings based on the context within a sentence of limited length, i.e., \emph{intra-sentence} contextual word embeddings.
Different from the naive BERT model, IsBERT learns the context-dependent word embeddings across different sentences, i.e.,  \emph{inter-sentence} contextual word embeddings. The implicit relations among tokens are captured, passed and strengthen across different sentences.

IsBERT mainly consists of an overlapped input processor and an inter-sentence encoder. Specifically, the input processor adds special tokens to better indicate boundaries and continuous in a long document. The input processor also overlaps each pair of adjacent sentences with a certain ratio of tokens to preserve inter-sentence context. To better capture the context information across different sentences in a long document, we further propose an inter-sentence encoder with the bidirectional overlapped contextual embedding learning and multi-hop inference mechanisms. The inter-sentence encoder contains multiple hop, where each hop learns the inter-sentence context-dependent word embeddings through a forward overlapped contextual embedding layer and a backward overlapped contextual embedding layer. The implicit relations among tokens, i.e., the inherent signals about whether certain words have been recognized in other context, are captured and passed across different sentences bidirectionally during learning. Different hops encode the documents repeatedly to update the embeddings and strengthen the inherent signals.

IsBERT takes full advantage of inter-sentence context in long documents, while for short documents with much less abundances of context, it may lose advantages. The predefined overlapping embeddings ratio, which works well for long documents, may not help capture and pass meaningful signals in a much shorter document, but lead to many repeated context and redundant embeddings in the model. To derive benefit from different documents with diverse abundances of context, we further propose an advanced \emph{Adaptive Inter-sentence BERT Model} (Ada-IsBERT). Ada-IsBERT has similar architecture as IsBERT, while it has an adaptive bidirectional overlapping mechanism instead of using a predefined overlapped embedding ratio. It dynamically adjust and update the inter-sentence overlapping ratio according to the length of different documents during training. With the adaptive overlapping mechanism, Ada-IsBERT is much robuster than IsBERT for documents with different abundances of context.

The main contributions of this paper are listed as follows.

$\bullet$ \textbf{C1}: We propose a fine-grained annotation scheme based on anthroponymy. This fine-grained annotation scheme provides information on various forms of person names (Section~\ref{Fine-grained_Annotations}).

$\bullet$ \textbf{C2}: We create the first dataset consisting of diverse academic homepages where the person names are fully annotated under our proposed fine-grained annotation scheme, called \textsl{FinegrainedName}, for the research of name recognition (Section~\ref{FinegrainedName_Dataset}).

$\bullet$ \textbf{C3}: We propose a \emph{Co-guided Neural Network} (CogNN) model to recognise person names using the fine-grained annotations. CogNN learns the different name form classes with two neural networks while fusing the learned signals through co-attention and gated fusion mechanisms (Section~\ref{CogNN_model}).

$\bullet$ \textbf{C4}: We also propose a \emph{Inter-sentence BERT Model} (IsBERT) to recognise fine-grained person names in long documents by capturing the inter-sentence context and implicit relations. IsBERT consists of an overlapped input processor, and an inter-sentence encoder with bidirectional overlapped contextual embedding learning and multi-hop inference mechanisms (Section~\ref{NameRec_model}).

$\bullet$ \textbf{C5}: We propose an \emph{Adaptive Inter-sentence BERT Model} (Ada-IsBERT), which is robust for documents with diverse abundances of context. Ada-IsBERT dynamically adjust and update the inter-sentence overlapping ratio according to the length of different documents during training (Section~\ref{Ada_NameRec_model}).

$\bullet$ \textbf{C6}: We empirically investigate the superiority of our proposed models on both academic homepages and newswire texts compared with state-of-the-art NER models. Experimental results also show that our annotations can be utilised in different ways to improve the recognition performance (Section~\ref{exp}).

This article is an extended version of our earlier conference paper~\cite{dai2020person}.
In the conference paper, we presented a fine-grained annotation scheme, a new dataset, and a person name recognition model which fully explores the rich training signals of name forms and intra-sentence context. In this journal extension, we extend the techniques for person names recognition in long documents by exploring the document-level inter-sentence context. To this end, we present the IsBERT model to utilize the document-level context and capture implicit relations. We propose an overlapped input processor, and an inter-sentence encoder with bidirectional overlapped contextual embedding learning and multi-hop inference mechanisms. In addition, we present an Ada-IsBERT model with an adaptive bidirectional overlapping mechanism to derive benefit from documents with diverse abundances of context. We also conduct more extensive experiments to demonstrate the effectiveness of our proposed methods for documents with different context.

The rest of the paper is organized as follows. Section~\ref{related} provides a review
of the related work. Section~\ref{dataset} introduces the Fine-grained Annotation Scheme and
our FinegrainedName dataset annotated under this scheme. Section~\ref{CogNN_model}
presents our CogNN model that takes advantages of the fine-grained annotations to recognise
person names. Section~\ref{NameRec_model} introduces the IsBERT model that utilizes the inter-sentence context in long documents to improve the performance of person name recognition. Section~\ref{Ada_NameRec_model} introduces the Ada-IsBERT model that adaptive to documents with diverse abundances of context. Section~\ref{exp} presents experiments in on academic homepages and news articles. Section~\ref{conclusion} concludes the paper.

\section{Related Work}\label{related}

\textbf{Named entity recognition (NER)} aims to identify proper names in text and classify them into different types, such as person, organisation, and location \cite{nadeau2007survey}. Neural NER models have shown excellent performance on long texts which follow strict syntactic rules, such as newswire and Wikipedia articles \cite{huang2015conll,chiu2016conll,ma2016conll}. However, these NER models are less attractive when applied to texts which may not have consistent and complete syntax \cite{li2015tweet,dugas2016tweet}. Recent studies consider user-generated short texts from social media platforms such as Twitter and Snapchat \cite{lu2018multimodal,moon2018multimodal}. However, there are few NER studies on free-form text with incomplete syntax including person names with various forms, such as academic homepages, academic resumes, articles in online forums and social media.

\textbf{BIO and BIEO} tagging schemes \cite{borthwick1998exploiting} are often used for named entity recognition in well-formed text, such as news articles in CoNLL-2003 \cite{sang2003introduction} and wikipedia articles in WiNER \cite{ghaddar2017winer}. However, such annotations for name spans cannot reflect the patterns of various name forms and brings challenges for recognising persons names in free-form text. To the best of our knowledge, no other work has utilised anthroponymy \cite{felecan2012name} and fine-grained annotations to help recognise person names.

\textbf{Information Extraction (IE)} studies on academic homepages and resumes usually treat the text content as a document, upon which traditional NER techniques are applied. For example, \cite{yiqing2018} use a Bi-LSTM-CRF based hierarchical model to extract all the publication strings from the text content of a given academic homepage.
\cite{dai2020joint} capture the relationship between publication strings and person names in academic homepages, and extract them simultaneously. This technique does not apply to our problem as we assume no pre-knowledge about the publication strings.

\textbf{Person names} are often recognised together with other entities, such as locations and organisations \cite{huang2015conll,ma2016conll,chiu2016conll,dugas2016tweet}.
\cite{packer2010extracting} focus on extracting name from noisy OCR data by combining rule based methods, the Maximum Entropy Markov Model, and the CRF model using a simple voting-based ensemble.
\cite{minkov2005extracting} extract person names from emails using CRF. They design email specific structural features and exploit in-document repetition to improve the extraction accuracy.
 \cite{aboaoga2013arabic} study person name recognition in Arabic using rule-based methods. To the best of our knowledge, no other work has taken into account name forms and uses deep learning based models.

\textbf{Multi-task learning }models, which train tasks in parallel and share representations between related tasks, have been proposed to handle many NLP tasks. \cite{caruana1997multitask} propose to share the hidden layers between tasks, while keeping several task-specific output layers. \cite{sogaard2016deep} jointly learn POS tagging, chunking and CCG supertagging by using successively deeper layers. \cite{ma2018joint} propose a model for sentiment analysis by jointly learn the character features and long distance dependencies through concatenation procedures. Rather than directly sharing the representations or concatenating the representations of different tasks, our co-attention and gated fusion mechanisms allow our model to co-guide the jointly trained tasks without overwhelmed by the misleading signals.

\textbf{Bidirectional Encoder Representations from Transformers} (BERT)~\cite{devlin2019bert} is a pre-trained language model. BERT has an encoder and decoder architecture, and learns to predict both the left and right context using a multi-layer transformer~\cite{vaswani2017attention}. Many extensions of BERT has been proposed to solve domain-specific problems, such as BioBERT~\cite{lee2020biobert} for biomedical text. There are also many studies try to compress the model size for real-life applications and on resource-restricted devices, such as TinyBERT~\cite{jiao2019tinybert}. However, these extensions have the same limitation on input length as the vanilla BERT. They can only capture the word embeddings based on intra-sentence context. Different from existing works, we explore the inter-sentence context, and propose methods based on BERT for documents with different abundances of context.

\section{Annotation Scheme and Dataset}\label{dataset}
We first present our fine-grained annotation scheme and introduce our FinegrainedName dataset annotated under this scheme.

\subsection{Fine-grained Annotations}\label{Fine-grained_Annotations}
Fine-grained annotations are done to better capture the person name form features in free-form texts. Annotating the name tokens with fine-grained forms offers more direct training signals to NER models to learn the patterns of person names.
Thus, unlike traditional NER datasets, which only label a name token with a \textit{PER} (person) label, we further provide fine-grained name form information for each name token based on anthroponymy \cite{felecan2012name}.

We label each name token using a three-dimensional annotation scheme:

\begin{itemize}
\item \textit{BIE}: \emph{Begin}, \emph{Inside}, or \emph{End} of name, indicating the position of a token in a person name,
\item \textit{FML}: \emph{First}, \emph{Middle}, or \emph{Last} name, indicating whether a name token is used as the first, middle, or last name, and
\item \textit{FI}: \emph{Full} or \emph{Initial}, indicating whether a name token is a full name word or an initial.
\end{itemize}

Using the three-dimensional annotation scheme above, we can describe the fine-grained name form of a name token.
For example, in Figure \ref{web_example}, `\texttt{John Doe}' can be labelled as \textit{Begin\_First\_Full End\_Last\_Full}, while `\texttt{Johnny van der Doe}' can be labelled as \textit{Begin\_First\_Full Inside\_Last\_Full Inside\_Last\_Full End\_Last\_Full}.

\subsection{The FinegrainedName Dataset}\label{FinegrainedName_Dataset}
FinegrainedName is a collection of academic homepages with person names fully annotated using the proposed annotation scheme. We use Selenium\footnote{\url{https://www.seleniumhq.org/}}, an open-source automated rendering software, to render the webpages and collect visible texts from the webpages. We download academic homepages from universities and research institutes around the world and focus on English homepages.

\begin{figure}[!t]
\includegraphics[width=\columnwidth]{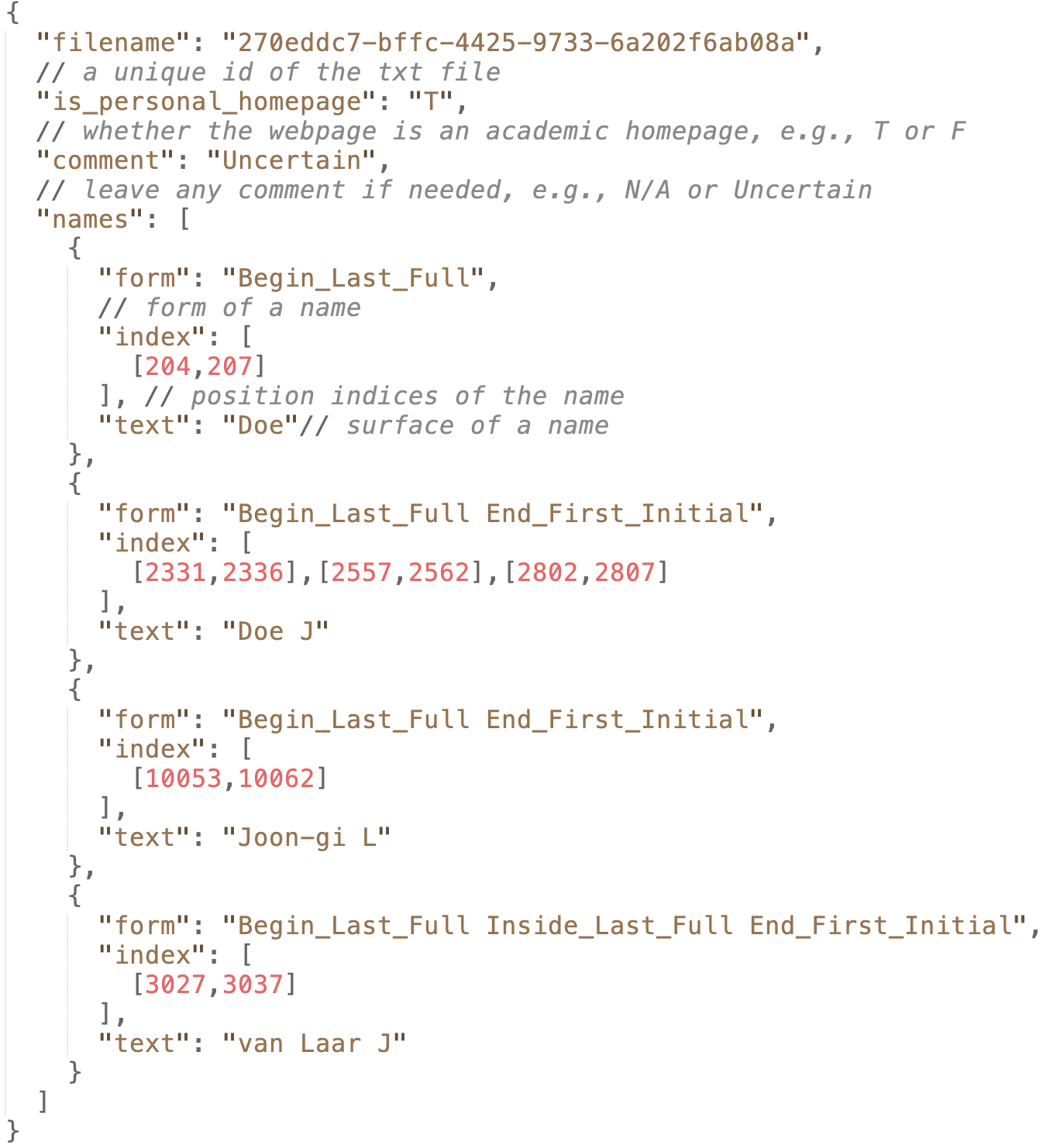}
\caption{Screenshot of an example JSON file}
  \label{json_example}
\end{figure}

FinegrainedName contains 2,087 subfolders and each subfolder contains three files for a webpage:
\begin{itemize}
\item An HTML file containing the page source.
\item A TXT file containing the visible text of the webpage, which is rendered by python's Selenium\footnote{\url{https://selenium-python.readthedocs.io/}} package.
\item A JSON file containing name annotations. Figure \ref{json_example} shows the example format of the JSON files.
\end{itemize}

\noindent\textbf{Annotation Tool }
Annotation of homepages is time-consuming, especially when a homepage contains many names in complex forms. We developed a semi-automatic tool to assist the annotation, which has five main functionalities:
\begin{itemize}
  \item \textit{Group\_label}: This functionality helps annotate a group of names of the same form. For example,  \texttt{`Doe J'} and \texttt{`Joon-gi L'} have the same forms and can be annotated at once.

  \item \textit{Index}: This functionality helps find all positions of a given name string in the TXT file.

  \item \textit{Mask}: This functionality helps annotators to proofread the text and find unlabelled names.
   It replaces all the names already annotated with a special token \textit{`ANNOTATED'}.

  \item \textit{Validate}: This functionality runs a simple automated quality check of the annotations. It checks: (1) whether the position indices of the names annotated in the JSON file are consistent with the names  appeared in the TXT file; and (2) whether each annotated name comes with the name form under the three-dimensional annotation schemes.

   \item \textit{Compare}: This functionality locates disagreement between two annotators' labels on the same homepage. It identifies the list of names with inter-annotator disagreement.

\end{itemize}

\noindent\textbf{Annotators }
There are 6 annotators to annotate the dataset.  The annotators are postgraduate students who have taken machine learning subjects. We provide a one-hour training to each annotator.

We provide the annotators with an annotation scheme and two example pages that are already annotated. We ask each annotator to annotate six pages. We examine the results and provide guidance on how to improve the annotation quality.

We highlighted the following at training:
\begin{itemize}
  \item Any named entities such as places, buildings, organizations, prizes, honored titles or books, which are named after a person, should not be annotated as a person's name.
  \item Words connected with a hyphen or an apostrophe should not be split into multiple tokens.  For example, both \texttt{`Joon-gi'} and \texttt{`O'Keeffe'} both have only one token.
  \item Nobiliary particles\footnote{\url{https://en.wikipedia.org/wiki/Nobiliary_particle/}}, e.g., \texttt{`van'}, \texttt{`zu'} and \texttt{`de'}, should be annotated as last names.
\end{itemize}

Each academic homepage is annotated by two annotators using our annotation tool. Any pages with uncertain name labels is noted down in the \texttt{comment} field (cf. Figure~\ref{json_example}). After their annotations, we make a decision on the disagreement between annotators and also check the uncertain pages and names. We send feedback when they annotate every 230 homepages.
ur

\noindent\textbf{Annotation Analysis }We summarise the disagreement between annotators.

\begin{itemize}

\item Confidence: Only 3.64\% of all the homepages contain annotations that are uncertain as flagged by the annotators, while 78.08\% of these pages are actually correctly labelled. This indicates that the annotators have high confidence in their annotations.

\item Inter-annotator Agreement: We compute the inter-annotator agreement on name strings and name forms using Cohen's Kappa measurement. The annotators have higher agreement on name strings ($\kappa = 0.63$) and lower agreement on fine-grained name forms ($\kappa = 0.41$). The disagreement is mainly in homepages with a long string of consecutive name tokens such that different annotators may disagree on which tokens to form a name. The annotators may also disagree on whether a name token is a first name, middle name, or last name. This is difficult especially when the context is unclear.
\item Time: On average, it takes 16 minutes to annotate an academic homepage with our tool.

\end{itemize}

\begin{table}[t!]
\begin{center}
\caption{\label{supplemental_tb} Summary of annotation and dataset. $\kappa$ is the Cohens Kappa measurement.}
\begin{tabular}{c  c c r}
\toprule

\multicolumn{4}{c}{\bf Summary of Annotation}
\\
\midrule
\multirow{2}{*}{Confidence} & \multicolumn{2}{c}{Uncertain pages } & 3.64\% \\
&\multicolumn{2}{c}{Accuracy on uncertain names}& 78.08\% \\
\midrule

\multirow{2}{*}{\shortstack{Inter-annotator\\ agreement ($\kappa$)}} & \multicolumn{2}{c}{Names} & $0.63$ \\
&\multicolumn{2}{c}{Name forms}& $0.41$\\
\midrule

Time & & &  16 min\\
\midrule
\midrule
\multicolumn{4}{c}{\bf Summary of Dataset}
\\ \midrule
\multicolumn{2}{c}{Total Homepages} & \multicolumn{2}{c}{2,087}\\
\multicolumn{2}{c}{Total Institutes} & \multicolumn{2}{c}{286}\\
\multicolumn{2}{c}{Average Institutes} & \multicolumn{2}{c}{7.29}\\
\multicolumn{2}{c}{STD. Institutes} & \multicolumn{2}{c}{7.27}\\
\midrule
\multicolumn{2}{c}{Total Names Indexes} & \multicolumn{2}{c}{70,864}\\
\multicolumn{2}{c}{Total Names} & \multicolumn{2}{c}{34,880}\\
\midrule
\multicolumn{2}{c}{Contain Initial} & \multicolumn{2}{c}{23,221}\\
\multicolumn{2}{c}{Begin with Last Name} & \multicolumn{2}{c}{22,581}\\
\multicolumn{2}{c}{Begin with Middle Name} & \multicolumn{2}{c}{13}\\
\multicolumn{2}{c}{Begin with First Name} & \multicolumn{2}{c}{12,286}\\
\bottomrule

\end{tabular}
\end{center}
\end{table}

\noindent\textbf{Dataset Analysis }
In total, the FinegrainedName contains 2,087 English academic homepages from 286 institutes, i.e., 7.29 pages per institute (standard deviation 7.27). A total of 34,880 names are annotated and 70,864 name
position indices are recorded. On average, a name appears twice in an academic homepage. Most names begin with last names (64.73\%) while the rest mostly begin with first names. Only 13 names start with middle names. Most names contain at least one initial (66.57\%). The two most frequent name forms are \textit{Begin\_Last\_Full End\_First\_Initial} and \textit{Begin\_First\_Full End\_Last\_Full}.
 Table \ref{supplemental_tb} summarises the annotation results and the dataset.

\begin{figure*}[!t]
\includegraphics[width=0.95\textwidth]{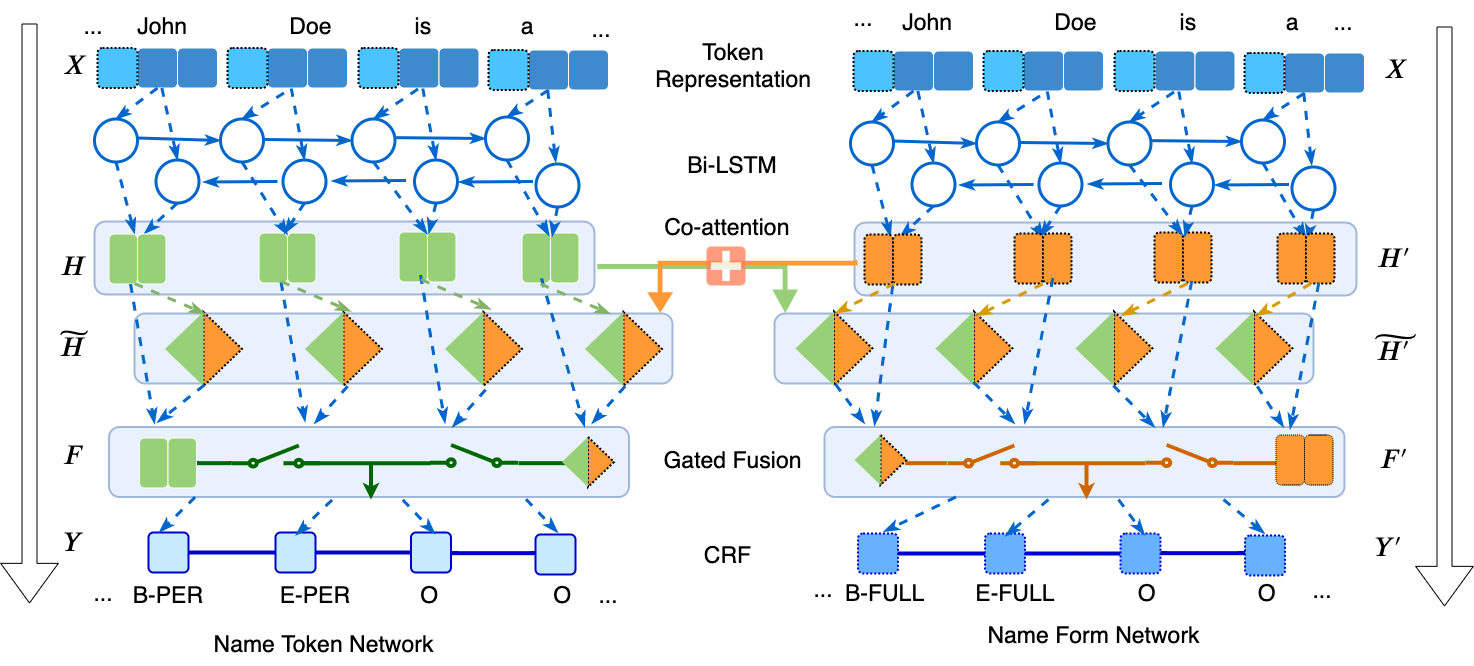}
\caption{CogNN network structure.} \label{overall}
\end{figure*}

\section{Proposed Models}
\subsection{CogNN Model}\label{CogNN_model}

The fine-grained annotations offer more direct training signals to NER models but also bring challenges because more label classes need to be learned. In this section, we present our CogNN model that takes advantages of the fine-grained annotations to recognise person names\footnote{A demonstration of our model will be available at \url{http://www.ruizhang.info/namerec/}}.

Given a sequence of input tokens $\boldsymbol{X}$, where $\boldsymbol{X} = [\boldsymbol{x_1}, \boldsymbol{x_2}, ..., \boldsymbol{x_n}]$ and $n$ is the length of the sequence, our aim is to predict for each token $\boldsymbol{x_i}$ whether it is a name token.\footnote{We use $\boldsymbol{x_i}$ to denote both a token and its embedding vector as long as the context is clear.}

Our proposed model CogNN achieves this aim with the help of two Bi-LSTM-CRF based sub-networks:  the \emph{name token network} and the \emph{name form network}, as illustrated in Figure~\ref{overall}.
The name token network focuses on predicting whether a token is part of a name (the \textit{BIE} dimension), while the name form network focuses on predicting the fine-grained name form class of the token (\textit{FML} or \textit{FI} dimensions). The intuition is that knowing whether a token is part of a name helps recognise the fine-grained name form class of the token, and vice versa. For example, if a token is not considered as part of a name, then even if it is a word initial, it should not be labelled as a name initial. To better capture the underlying correlation between different annotations, we share the learning signals of two sub-neural networks through a co-attention layer. To avoid being overwhelmed by possible misleading signals, we further add a gated-fusion layer to balance the information learned from two sub-networks.

In particular, an input token is represented by concatenating its word embedding and its letter case vector. We feed such representation of the input into Bi-LSTM to learn its hidden representation matrix, which is detailed in Section \ref{sub_lstm}.
Then, we use co-attention and gated fusion mechanisms to co-guide the two jointly trained sub-networks. Our co-attention mechanism updates the importance of each token learned from the two sub-networks and records their correlations (Section \ref{sub_attention}). Our gated fusion mechanism helps decide whether and how much to accept new signals from the other sub-network (Section \ref{sub_gate}). The two sub-networks are trained simultaneously by minimising their total loss (Section \ref{joint_train}).

\subsubsection{Capture: Hidden Feature Extraction} \label{sub_lstm}

The name token network (denoted as $N_Y$)
and the name form network (denoted as $N_{Y^{\prime}}$) have a similar structure. They only differ in the target labels $\boldsymbol{Y}$ and $\boldsymbol{Y^{\prime}}$.
Here, $\boldsymbol{Y}$ denotes the label sequence that indicates whether an input token is part of a name,
and $\boldsymbol{Y^{\prime}}$ denotes the label sequence
that indicates the form class of each input token.
We focus our explanation on the name token network $N_Y$ in the following discussion while the name form network works in a similar way.

An input token $\boldsymbol{x_i} \in \boldsymbol{X}$ is represented by concatenating its word embedding $\boldsymbol{e_i}$ and its letter case vector $\boldsymbol{s_i}$.
We use GloVe \cite{pennington2014glove} computed on our FinegrainedName corpus for the word embeddings $\boldsymbol{e_i}$.
The letter case vector $\boldsymbol{s_i}$ indicates the letter case  information of $\boldsymbol{x_i}$, which is an important hint for recognising names.
For example, the first letter of a name token is often in uppercase, and a name initial is often formed by an uppercase letter plus a dot.
Our letter case vector  is a three-dimensional binary vector where each dimension represents:
(i) whether the first character in the token is in uppercase, (ii) whether all  characters in the token are in uppercase, and (iii) whether any character in the token is in uppercase.

We then use Bi-LSTM \cite{dyer2015transition} to capture the hidden features from the input sequence. The output hidden representation, denoted as $\boldsymbol{h_i}$,
 summarises the context information of $\boldsymbol{x_i}$ in $\boldsymbol{X}$. Our hidden representation matrix $\boldsymbol{H}$ in  $N_Y$ can be written as $[\boldsymbol{h_1},\boldsymbol{h_2},...,\boldsymbol{h_n}]$, where $\boldsymbol{h_i}  \in \mathcal{R}^d$ and $d$ is the number of dimensions of the hidden representation.
 Similarly, $\boldsymbol{H'}$ in $N_{Y^{\prime}}$ can be written as $[\boldsymbol{h_1^{\prime}},\boldsymbol{h_2^{\prime}},...,\boldsymbol{h_n^{\prime}}]$.

\subsubsection{Share: Co-attention Mechanism} \label{sub_attention}

Training the two sub-networks separately is suboptimal, since the
underlying correlation among the name label dimensions is lost.
For example, a token recognised as \emph{Inside} in $N_Y$ is more possible to be \emph{Middle}  in $N_{Y^{\prime}}$. To address this issue, we share the learning signals between the hidden representation matrices $\boldsymbol{H}$ and $\boldsymbol{H^{\prime}}$, and obtain new hidden representation matrices $\boldsymbol{\tilde{H}}$ and $\boldsymbol{\tilde{H^{\prime}}}$ for the two sub-networks, respectively.

Specifically, we use co-attention to take the learning signals from two hidden representations into account by:
\begin{align*}
\boldsymbol{P} &= tanh(W_h \boldsymbol{H}  \oplus ( W_{h^{\prime}} \boldsymbol{H^{\prime}}  +  b_{h^{\prime}} ))
\end{align*}
\noindent where $W_h$ and $W_{h^{\prime}} \in \mathcal{R}^{k\times d}$ are trainable parameters, $k$ is dimensionality of the parameters,
$\oplus$ is the concatenating operation, $tanh$ is the activation function to scale into the range of (-1,1), and $P \in \mathcal{R}^{2k\times n}$.

The co-attention distribution that records the importance of each token after examining two hidden representation sequences can be obtained as:
\begin{align*}
\boldsymbol{A} &= softmax(W_{p} \boldsymbol{P}  +  b_{p})
\end{align*}
\noindent where $W_{p} \in \mathcal{R}^{1\times 2k}$ are trainable parameters and $A \in \mathcal{R}^{n}$ is an importance weight matrix.

The new hidden representation $\boldsymbol{\tilde{h_i}}$ can be computed by:
\begin{align*}
\boldsymbol{\tilde{h_i}} &= \boldsymbol{a_i} \boldsymbol{h_i},  \boldsymbol{a_i} \in \boldsymbol{A}, \boldsymbol{h_i} \in \boldsymbol{H}
\end{align*}
We thus obtain the new hidden representation sequences  $\boldsymbol{\tilde{H}}=[\boldsymbol{\tilde{h_1}},\boldsymbol{\tilde{h_2}},...,\boldsymbol{\tilde{h_n}}]$
and $\boldsymbol{\tilde{H^{\prime}}} = [\boldsymbol{\tilde{h_1^{\prime}}},\boldsymbol{\tilde{h_2^{\prime}}},...,\boldsymbol{\tilde{h_n^{\prime}}}]$ for the two sub-networks.

\subsubsection{Balance: Gated Fusion Mechanism} \label{sub_gate}
To avoid being overwhelmed by misleading learning signals from the other sub-network $N_{Y^{\prime}}$ during co-attention,
we dynamically balance the information learned from the (independent) hidden representation $\boldsymbol{H}$
and the corresponding new (dependent) hidden representation $\boldsymbol{\tilde{H}}$ for $N_Y$ (also $\boldsymbol{H'}$ and $\boldsymbol{\tilde{H'}}$ for $N_{Y^{\prime}}$), and  obtain a fused representation matrix $\boldsymbol{F}$ ($\boldsymbol{F'}$ for $N_{Y^{\prime}}$).

Inspired by the study on multi-modal fusion for images and text \cite{lu2018multimodal}, we add a gated fusion layer to balance the information from
$\boldsymbol{H}$ and $\boldsymbol{\tilde{H}}$  to obtain better representations.

We first transform each item in $\boldsymbol{H}$ and $\boldsymbol{\tilde{H}}$ by:
\begin{align*}
\boldsymbol{h_{\tilde{h_i}}} &= \tanh(W_{\tilde{h_i}} \boldsymbol{\tilde{h_i}} +  b_{\tilde{h_i}})\\
\boldsymbol{h_{h_i}} &= \tanh(W_{h_i} \boldsymbol{h_i} +  b_{h_i})
\end{align*}
\noindent where $W_{\tilde{h_i}}$ and $W_{h_i}$ are trainable parameters.

Then, our fusion gate, which decides whether and how much to accept the new information, is computed as:
\begin{align*}
\boldsymbol{g_t} &= \sigma(W_{g_t} (\boldsymbol{h_{\tilde{h_i}}}  \oplus  \boldsymbol{h_{h_i}}))
\end{align*}
\noindent where $\sigma$ is the element-wise sigmoid function to scale into the range of (0,1) and $W_{g_t}$ are trainable parameters.

We fuse the two representations using the fusion gate through:
\begin{align*}
\boldsymbol{f_i} &=  \boldsymbol{g_t} \boldsymbol{h_{h_i}} + (1-\boldsymbol{g_t}) \boldsymbol{h_{\tilde{h_i}}}
\end{align*}
The fused representation sequence $\boldsymbol{F}=[\boldsymbol{f_1},\boldsymbol{f_2},...,\boldsymbol{f_n}]$ is trained to produce a label sequence $\boldsymbol{Y}$. To enforce the structural correlations between labels, $\boldsymbol{Y}$ is passed to a CRF layer to learn the correlations of the labels in neighborhood. Let $\mathcal{Y}$
denotes the set of all possible label sequences for $\boldsymbol{F}$. Then, the the probability of the label sequence $\boldsymbol{Y}$ for a given fused representation sequence $\boldsymbol{F}$ can be written as :
\begin{align*}
p(\boldsymbol{Y}|\boldsymbol{F},W_{Y}) &= \frac{\prod_{t} \psi_{t} (y_{t-1}, y_t; \boldsymbol{F})} {\sum_{Y^\prime \in \mathcal{Y}} {\prod_{t} \psi_{t} ({y^\prime}_{t-1}, {y^\prime}_t; \boldsymbol{F})}}
\end{align*}
\noindent where $\psi_{t} (y^\prime, y; \boldsymbol{F}) $ is a potential function, $W_{Y}$ is a set of parameters that defines the weight vector and bias corresponding to label pair $(y^\prime, y)$.

Similarly,  we can also compute the fused representation sequence $\boldsymbol{F^\prime}$ and $p(\boldsymbol{Y^{\prime}}|\boldsymbol{F^\prime},W_{Y^{\prime}})$.

\subsubsection{Joint Training} \label{joint_train}
The remaining question is how to train two networks simultaneously to produce label sequences $\boldsymbol{Y}$ and $\boldsymbol{Y^{\prime}}$. We achieve this by joint optimisation. Specifically, we train the CogNN model end-to-end by minimising loss $\mathcal{L}$, which is the sum of the loss of the two sub-networks:
\begin{align*}
\mathcal{L} &= \mathcal{L}(W_Y)+\mathcal{L}(W_{Y^{\prime}})
\end{align*}

\noindent where $\mathcal{L}(W_Y)$ and $\mathcal{L}(W_Y)$ are the negative log-likelihood of the ground truth label sequences $\boldsymbol{\hat{Y}}$ and $\boldsymbol{\hat{Y^{\prime}}}$ for the input sequences respectively, which are computed by:

\begin{small}
\begin{align*}
\mathcal{L}(W_Y) &= - \sum_i \sum_{\boldsymbol{Y_i}} \delta{(\boldsymbol{Y_i} = \boldsymbol{\hat{Y}}})\log p(\boldsymbol{Y_i}|\boldsymbol{F})
\\
\mathcal{L}(W_{Y^{\prime}}) &= - \sum_j \sum_{\boldsymbol{Y^{\prime}_j}} \delta{(\boldsymbol{Y^{\prime}_j} = \boldsymbol{\hat{Y^{\prime}}})}\log p(\boldsymbol{Y^{\prime}_j}|\boldsymbol{F^\prime})
\end{align*}
\end{small}

\subsection{IsBERT Model}\label{NameRec_model}
CogNN fully explores the intra-sentence context and rich training signals of name forms. However, the inter-sentence context and implicit relations, which are extremely essential for recognizing person names in long documents, are not captured. In this section, we present our
IsBERT that takes advantages of inter-sentence context informations.

Given a sequence of sentences $D=[T_1, T_2, T_3,...,T_M]$ from a document $D$, where $T_j = [t_1, t_2, t_3,..,t_n]$, $n$ is the length of the sentence and $t_i$ is the $i$-th tokens in $T_j$, we aim to predict whether $t_i$ is a name token and the corresponding fine-grained name type.

Our IsBERT achieves this aim by learning inter-sentence contextual word embeddings. The context information across different sentences are shared in a long document through bidirectional overlapped embedding learning. The implicit relations among tokens, i.e., the inherent signals about whether certain words have been recognized in other context, are captured, passed and strengthen across different sentences during bidirectional overlapped embedding learning and multi-hop inferences. The intuition behind this is that some person name entities may appear repeatedly in different context, while some context provide strong signals about the entity type and others not. Passing the inter-sentence context information can help recognize more person names in different context.

\begin{figure}[t!]
  \centering
  \includegraphics[width=\columnwidth]{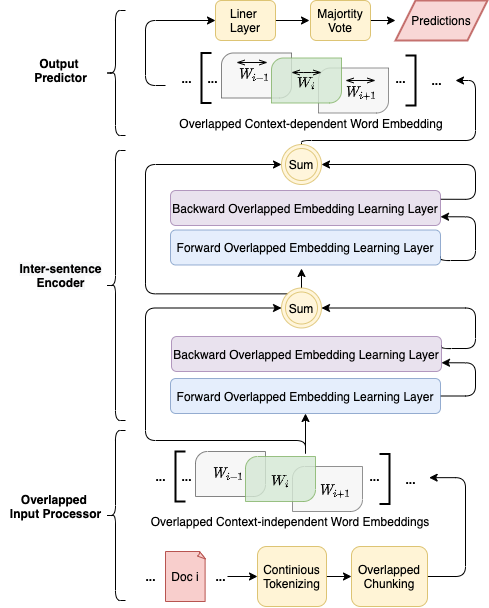}
  \caption{The overall architecture of the IsBERT model.}
  \label{fig:docbert_structure}
\end{figure}

Figure~\ref{fig:docbert_structure} illustrates the overall structure of the IsBERT model. The model consists of three parts: overlapped input processor, inter-sentence encoder and output predictor. The input processor converts input documents into chunks with overlapped \emph{context-independent} embeddings (Section~\ref{sec:Pre-processing}).
The inter-sentence encoder contains multi-hop inference modules, where each inference module learns the inter-sentence \emph{context-dependent} word embeddings through overlapped contextual embedding learning, and different inference modules encode the documents forward and backward alternately to update the embeddings and pass the inherent signals across sentences (Section~\ref{sec:Overlapped Embedding Mechanism}).
Finally, the output predictor produce the final predictions for each input token based on the learned inter-sentence contextual word embeddings (Section~\ref{sec:Output Predictor}).

\begin{figure*}[t!]
  \centering
  \includegraphics[width=0.8\textwidth]{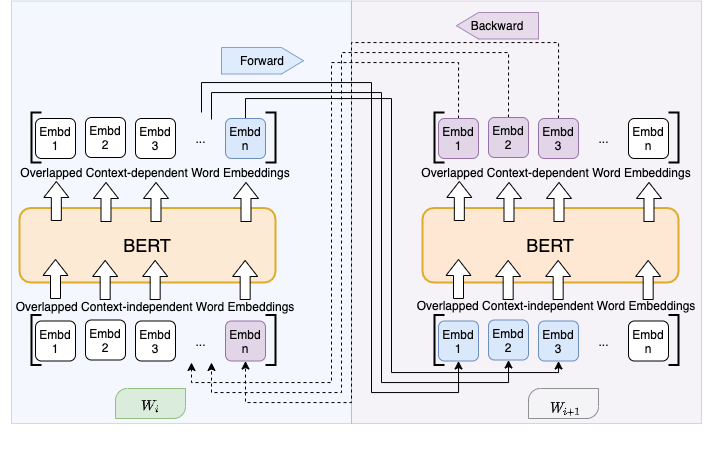}
  \caption{The forward and backward overlapped embedding learning in IsBERT.}
  \label{fig:docbert_structure_detail}
\end{figure*}

\subsubsection{Overlapped Input Processor}
\label{sec:Pre-processing}

We focus on English documents. The documents are first tokenized by whitespace and punctuations followed by WordPiece tokenization~\cite{wu2016google}. While whitespace tokenizer breaks the input documents into tokens around the whitespace boundary, the WordPiece tokenizer uses longest prefix match to further break the tokens into sub-tokens. The resultant sub-tokens which from one token are assigned the same label.
For example, if the token ``\verb\zhenning\'' with label \textit{Begin\_First\_Full} is tokenized into ``\verb\z\'', ``\verb\##hen\'' and ``\verb\##ning\'', then all of the three sub-tokens are labelled as \textit{Begin\_First\_Full}. For convenience's sake, we will use ``token'' to refer to ``resultant sub-tokens from WordPiece tokenizer'' in the following discussion.

After tokenizing the documents into a series of tokens, the most straightforward way to learns contextual word embeddings is to split the documents into different chunks sentence by sentence and feed them into the naive BERT model~\cite{devlin2019bert}. The sentences exceeding the input length limit are broken into several short sub-sentences before fed into the model.

Such a straightforward method only capture the \emph{intra-sentence} context since each input to the model is a single distinct sentence, while the \emph{inter-sentence} context and relationships are lost during learning. For applications which assume the independence between different sentences, only capturing the intra-sentence context is acceptable. However, for applications which strongly rely on document-level information, capturing the inter-sentence context is important.

To address this issue, we propose an overlapped chunking method to preserve inter-sentence information when learning the contextual word embedding using BERT. Two strategies are used in the overlapped chunking method compared with a naive pre-processing method in BERT: 1) adding special tokens to indicate the boundaries and continuous in document; 2) overlapping each pair of adjacent chunks with predefined ratio.

\begin{figure}[t!]
  \centering
  \includegraphics[width=\columnwidth]{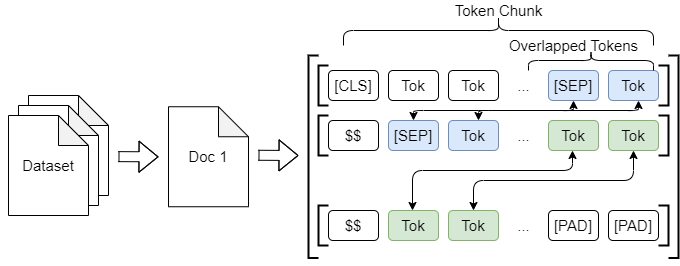}
  \caption{An example of converting a document to token chunks}
  \label{fig:doc_to_token_chunks}
\end{figure}

Specifically, ``[CLS]'' is added at the beginning of the first chunk of each document to indicate the document-level boundary, while the remaining chunks of the document are started with the special token ``\$\$'' to indicate the continuation. The special token ``[SEP]'' is appended to the end of each sentence to specify the sentence-level boundary. In addition to adding signals to indicate the boundaries and continuous, each pair of adjacent chunks from one document is overlapped with a certain ratio of tokens, i.e., last $k$ tokens of one chunk is the first $k$ tokens of the follow-up chunk. An example of the pre-processed overlapped input documents is shown in Figure~\ref{fig:doc_to_token_chunks}.

For sentence-level applications, different chunks need to be shuffled in the sentence-level after tokenization to avoid over-fitting during training. In our overlapped input processor, the shuffling only happens in the document-level, while the order of different chunks from each document is maintained. Specifically, given a dataset $U$ containing $L$ documents $U=[D_1, D_2, D_3,...D_L]$, where each document is pre-processed and splited into a list of chunks $D_l=[T_1, T_2, T_3,...,T_M]$. Then, after the shuffling of the dataset, only the order of $D_l$ in $U$ should be changed, while the order of $T_j$ in each $D_l$ remains as before. Then, the tokens in each chunk are mapped to corresponding \emph{context-independent} word embeddings and each document $D$ can be represented as $[W_1, W_2, W_3,...,W_M]$, where $W_j$ represents all the context-independent word embeddings of all the tokens in chunk $T_j$.

\subsubsection{Inter-sentence Encoder}
\label{sec:Overlapped Embedding Mechanism}

The input processor converts a document $D$ into chunks with overlapped \emph{context-independent} word embeddings. To better learn the inter-sentence \emph{context-dependent} word embeddings, we further propose an inter-sentence encoder. The chunks with context-independent word embeddings $[W_1, W_2, W_3,...,W_M]$ from one document are fed into the inter-sentence encoder in order. The inter-sentence encoder contains multi-hop inference modules, where each inference module learns the inter-sentence context-dependent word embeddings through two overlapped embedding learning layers, i.e., a forward overlapped embedding learning layer and a backward overlapped embedding learning layer. Figure~\ref{fig:docbert_structure_detail} illustrates the procedures or forward and backward overlapped embedding learning in IsBERT.

In the forward overlapped embedding learning layer, for each adjacent chunk pairs $W_{i-1}$ and $W_{i}$ in a document, we take the output context-dependent word embeddings of the overlapped tokens from $W_{i-1}$ as the input word embeddings of same tokens in $W_{i}$. Then, the corresponding forward context-dependent embedding  $\overrightarrow{W_i}$ is represented as:
\begin{equation}
    \overrightarrow{W_i} =BERT([\overrightarrow{W_{i-1}}] {\bot k}, W_i)
\end{equation}
where $[\overrightarrow{W_{i-1}}] {\bot k}$ is the last $k$ numbers of the context-dependent embeddings of $\overrightarrow{W_{i-1}}$.

In the backward overlapped embedding learning layer, for each adjacent chunk pairs $W_{i}$ and $W_{i+1}$ in a document, we feed the output context-dependent word embeddings of the overlapped tokens from $W_{i+1}$ back to $\overrightarrow{W_i}$.

Then, the forward and backward (inter-sentence) context-dependent embedding  $\overleftrightarrow{W_i}$, which carry the context information from both previous and succeeding chunks, is represented as:
\begin{equation}
    \overleftrightarrow{W_i} =BERT(\overrightarrow{W_i}, [\overleftarrow{W_{i+1}}] {\top k})
\end{equation}
where $[\overleftarrow{W_{i+1}}] {\top k}$ is the first $k$ numbers of the context-dependent embeddings of $\overleftarrow{W_{i+1}}$.


To strengthen the inter-sentence relations and strengthen the inherent signals in the context-dependent embeddings, we apply multi-hop inference modules and sum the learned inter-sentence context dependent embedding $[\overleftrightarrow{W_1}, \overleftrightarrow{W_2}, ...,\overleftrightarrow{W_M}]_t$ of the current inference module $\boldsymbol{I}_t$ with the inter-sentence context-dependent embeddings $[\overleftrightarrow{W_1}, \overleftrightarrow{W_2}, ...,\overleftrightarrow{W_M}]_{t-1}$ from the previous inference module $\boldsymbol{I}_{t-1}$ as the input for the next inference module $\boldsymbol{I}_{t+1}$.
Then, after multi-hop inferences, the last inter-sentence context-dependent embeddings can be represented as $[\overleftrightarrow{W_1}, \overleftrightarrow{W_2}, ...,\overleftrightarrow{W_M}]_m$, where $m$ is the inference times.

\subsubsection{Output Predictor}
\label{sec:Output Predictor}
The inter-sentence encoder converts chunks with overlapped \emph{context-independent} word embeddings $[W_1, W_2, W_3,...,W_M]$ into inter-sentence \emph{context-dependent} word embeddings $[\overleftrightarrow{W_1}, \overleftrightarrow{W_2}, ...,\overleftrightarrow{W_M}]_m$. Then, the learned inter-sentence contextual word embeddings are passed through a linear layer to get final prediction. The predictions for wordpieces (sub-tokens) from the same token are inconsistent. For these cases, we apply the majority vote and random tiebreaker to decide the final predictions for these token.

\subsection{Ada-IsBERT Model}\label{Ada_NameRec_model}

IsBERT takes full advantage of inter-sentence context in long documents, while for short documents with much less abundances of context, it may lose advantages. The predefined overlapping ratio $k$, which works well for long documents, may not help capture and pass meaningful signals in a much shorter document. To derives benefit from different documents with diverse abundances of context, we introduce our Ada-IsBERT model in this section.

Ada-IsBERT has similar architecture as IsBERT, while it has an adaptive overlapped embedding ratio $\tilde k$ instead of using a predefined overlapped embedding ratio $k$. It dynamically choose the inter-sentence overlapping ratio from a list of ratios according to the length $l$ of different documents $[D_1, D_2, D_3,...D_L]$ during training and a list of predefined length thresholds. Then, the inter-sentence contextual embedding  $\overleftrightarrow{W_i}$ is represented as:
\begin{equation}
    \overleftrightarrow{W_i} =BERT(\overrightarrow{W_i}, [\overleftarrow{W_{i+1}}] {\top \tilde k})
\end{equation}
where $[\overleftarrow{W_{i+1}}] {\top \tilde k}$ is the adaptive number of the context-dependent embeddings of $\overleftarrow{W_{i+1}}$.

With the adaptive overlapping mechanism, Ada-IsBERT is much robuster than IsBERT for documents with different abundances of context.

\section{Experimental Study}\label{exp}

We explore the following five aspects of our approach by a comprehensive experimental study:
\begin{itemize}
\item The impact of using fine-grained annotations in different ways for recognising person names from academic homepages (Section~\ref{subsub_scheme}).
\item The performance of the proposed models against baseline models and variant models on recognising person names from academic homepages (Section~\ref{subsub_model}).
\item The impact of using different overlapped ratios and different numbers of inference module for recognising person names from academic homepages with different abundances of context (Section~\ref{subsub_para}).
\item The applicability of the proposed models on recognising person names from news articles (Section~\ref{exp_conll}).
\item The cases study and error analysis of our proposed models (Section~\ref{exp_case}).
\end{itemize}

\subsection{Effectiveness on Academic Homepages}\label{exp_academic}

In this subsection, we study the performance of our proposed annotation scheme together with the CogNN model on academic homepages.

\begin{figure}[t!]
  \centering
  \includegraphics[width=\columnwidth]{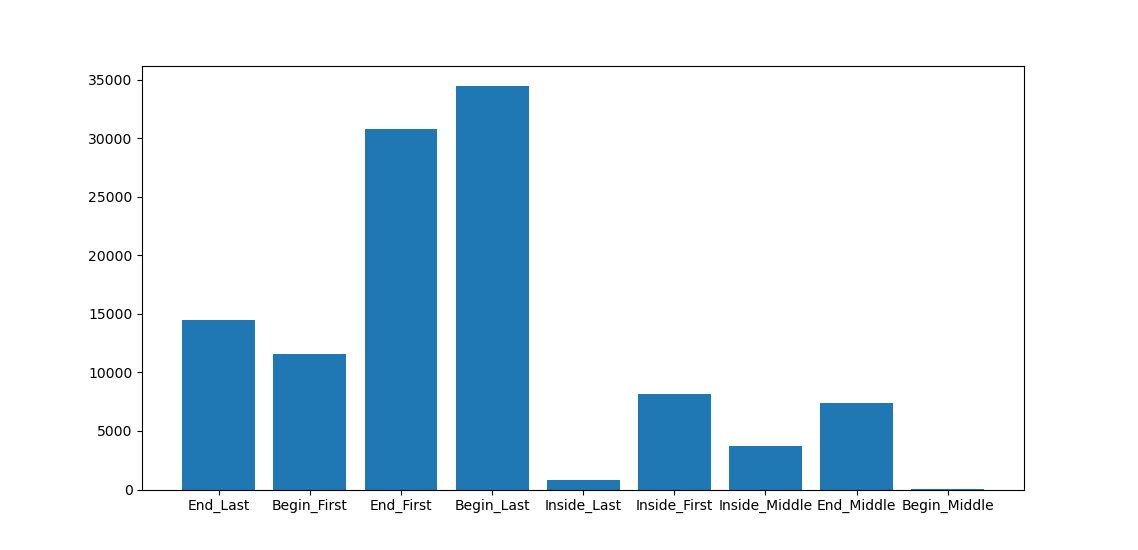}
  \caption{Training Dataset Label Distribution}
  \label{fig:train_dataset_dis}
\end{figure}

\begin{figure}[t!]
  \centering
  \includegraphics[width=\columnwidth]{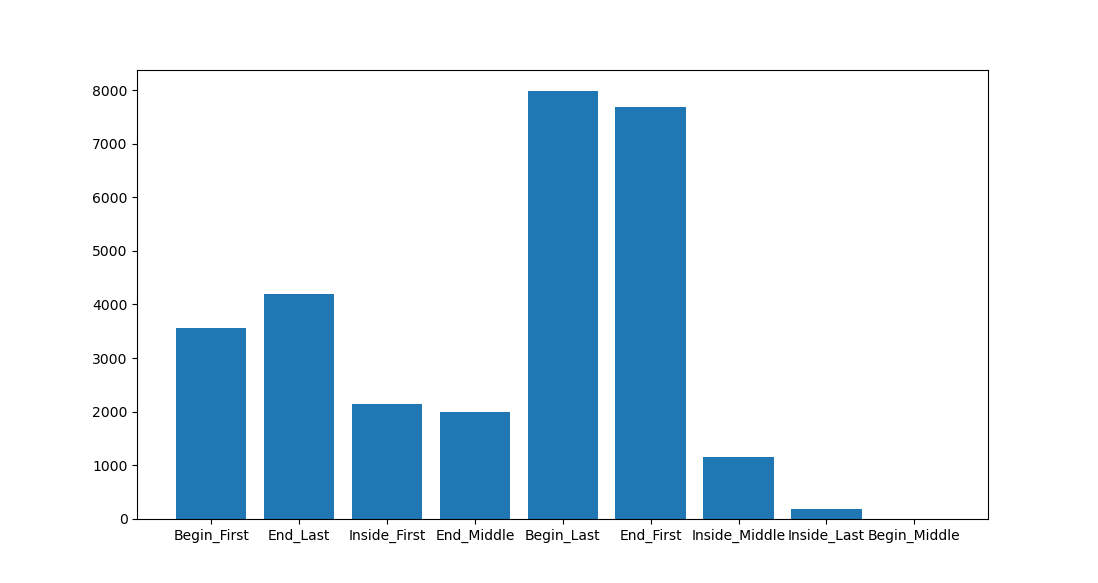}
  \caption{Testing Dataset Label Distribution}
  \label{fig:test_dataset_dis}
\end{figure}

\noindent\textbf{Dataset }
We use the FinegrainedName with the proposed fine-grained annotation scheme (Section \ref{dataset}), where 1,677 homepages are used for training and developing and 410 homepages are used for testing. The label distribution in two datasets are shown in Figure~\ref{fig:train_dataset_dis} and~\ref{fig:test_dataset_dis} respectively.

\noindent\textbf{Evaluation }
Recall (\textbf{R}), Precision (\textbf{P}) and F1-scores (\textbf{F}) are used to measure the performance. We report the \emph{Token Level} performance, which reflects the model capability to recognise each person name token. We also report the \emph{Name Level} performance, which reflects the model capability to recognise a whole person name without missing any token. The reported improvements are statistically significant with p $<$ 0.05 as calculated using McNemar's test.


\noindent\textbf{Word Embedding }
For CogNN model, we use GloVe\footnote{\url{https://nlp.stanford.edu/projects/glove/}} to learn word embeddings. For experiments on academic homepages, we train 100-dimensional word embeddings using GloVe on FinegrainedName, with a window size of 15, minimun vocabulary count of 5, full passes through cooccurrence matrix of 15, and an initial learning rate of 0.05. For experiments on newswire articles, we initialise word embeddings with GloVe pretrained 100-dimensional embeddings which are pretrained on English Gigaword Fifth Edition\footnote{\url{https://catalog.ldc.upenn.edu/LDC2011T07/}} containing a comprehensive archive of newswire text data.

\subsubsection{Effectiveness of the Annotation Scheme }\label{subsub_scheme}

\begin{figure}[!t]
\includegraphics[width=\columnwidth]{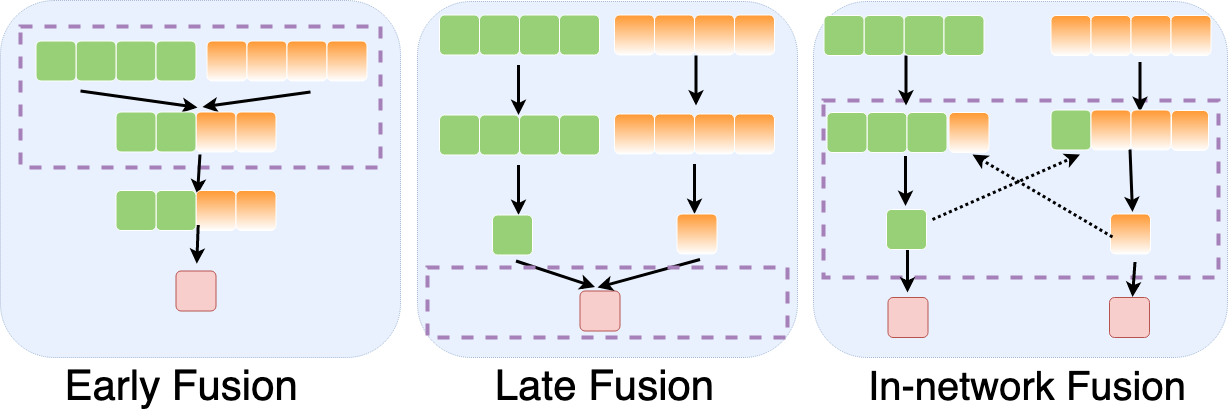}
\caption{Early, late, and in-network fusion.} \label{Early_Late_In}
\end{figure}

We first study the impact of using our fine-grained annotations with different fusion strategies (Figure~\ref{Early_Late_In}) and different models. Specifically, four fusion strategies are tested:
\begin{itemize}
\itemsep0em
\item \textbf{No fusion}: Training an independent model that learns to label the input sequence with the BIE, FML, or FI label types but not a combination of any two types of the labels.
\item \textbf{Early fusion}: Training an independent model that learns to label the input sequence with a cartesian product of the BIE, FML, and FI label types, e.g., to label `\texttt{John Doe}' with \textit{Begin\_First\_Full End\_Last\_Full}.
\item  \textbf{Late fusion}: Training sub-models each focusing on one label type and merging all the predicted labels afterwards to yield the final prediction by using every span of
tokens with name label as a name.
\item  \textbf{In-network fusion}: Training two sub-models each focusing on one label type and sharing the learning signals in the intermediate levels of the sub-models (This is what our proposed CogNN model does).
\end{itemize}

Three models are tested:
\begin{itemize}
\item \textbf{CRF}: \cite{finkel2005crf}. We use the Java implementation provided by the Stanford NLP group\footnote{\url{https://nlp.stanford.edu/software/CRF-NER.html}}. The software provides a generic implementation of linear chain CRF model.
\item \textbf{Bi-LSTM-CRF}:\cite{huang2015conll}. Specifically, the word embeddings are fed into a Bi-LSTM layer as input. Dropout is applied to the output of Bi-LSTM layer to avoid overfitting. The output is further fed into a linear chain CRF layer to predict the tokens labels. Specifically, it has a 100-dimensional hidden layer, a dropout layer with probability 0.5, a batch size of 32, and an initial learning rate of 0.01 with a decay rate of 0.05.
\item \textbf{CogNN}: Our proposed model is implemented following the description in Section \ref{CogNN_model}. Dropout is applied on the Bi-LSTM layers. We use a standard grid search to find the best hyperparameter values on the developing dataset. We choose the initial learning rate among $[0.001, \bf{0.01}, 0.1]$, the decay rate among $[\bf{0.05}, 0.1]$, the dimension of hidden layer among $[50, \bf{100}, 200]$, the dropout rate among $[0.2, \bf{0.5}]$ and has a batch size of 32. The optimal hyperparameters are highlighted above in bold.
\end{itemize}

All above deep learning models are implemented using Theano\footnote{\url{http://deeplearning.net/software/theano/}} and Lasagne\footnote{\url{https://lasagne.readthedocs.io/}}. All the optimal hyperparameters of above deep learning models are obtained with standard grid search on the same developing dataset with stochastic gradient descent. We stop training if the accuracy does not improve in 10 epochs.

Table \ref{result_scheme} shows the results.
CRF achieves an up to 13.83\% improvement on F1 score when the fine-grained FML and FI annotations are provided without any fusion. The same trend can also be observed for Bi-LSTM-CRF. The reason is that academic homepages contain person names with various forms and simple BIE annotations cannot reflect the patterns of name forms well. When examining the performance of using early fusion, we find that it is much worse than no fusion. This is expected as early fusion of different dimensions of name form leads to too many classes to be predicted. Even for a two-token name, it may have $(3 \times 3 \times 2)^2=324$ possible name form combinations. Late fusion offers better performance than no fusion with up to 15.95\% and 2.56\% improvements in F1 score for Bi-LSTM-CRF and CRF, respectively, which indicate that the separately trained models on different annotations have their own focuses. However, the underlying relationships among different name forms are not captured using late fusion. Our CogNN model, which uses in-network fusion, outperforms the best late fusion baseline by up to 5.35\% in F1-score. The reason is that our model can take advantage of the correlations between name form types when training and gain higher prediction confidence.

Overall, the fine-grained annotations can improve the performance of person name recognition on academic homepages using no fusion, late fusion and in-network fusion strategies. The neural-based models perform better than non-neural models and the in-network fusion can achieve the best results.


\begin{table}[t!]
\begin{center}
\caption{Name level F1 score of using the fine-grained annotations in different ways with different models on FinegrainedName.}\label{result_scheme}
\begin{tabular}{c|c|c|c}
\toprule
\parbox{1.6cm}{\centering \bf{Fusion\\Strategies}} & \parbox{1cm}{\centering \bf Models} & \bf Annotations & \bf F \\
\midrule
\multirow{6}{*}{\parbox{1cm}{\centering No\\Fusion}} & \multirow{3}{*}{CRF} & BIE & 41.15\\
& &  FML & 54.98\\
& &  FI & 50.32\\
\cmidrule(r){2-4}
&\multirow{3}{*}{ Bi-LSTM-CRF}&  BIE & 80.89\\
&&  FML & 82.11\\
&&  FI & 81.71\\
\midrule

\multirow{2}{*}{\parbox{1cm}{\centering Early\\Fusion}} &  CRF & BIE $\times$ FML $\times$ FI & 28.14\\
\cmidrule(r){2-4}
 &  Bi-LSTM-CRF & BIE $\times$ FML $\times$ FI & 62.65\\
\midrule

\multirow{8}{*}{\parbox{1cm}{\centering Late\\Fusion}} & \multirow{4}{*}{CRF}& BIE $\cup$ FML & 56.23\\
&  & BIE $\cup$ FI & 56.01\\
&  & FML $\cup$ FI & 56.38\\
&  & BIE $\cup$ FML $\cup$ FI & 57.10\\
\cmidrule(r){2-4}

& \multirow{4}{*}{Bi-LSTM-CRF}& BIE $\cup$ FML & 83.12\\
&  & BIE $\cup$ FI & 83.08\\
&  & FML $\cup$ FI & 83.29\\
&  & BIE $\cup$ FML $\cup$ FI & 83.45\\
\midrule

\multirow{4}{*}{\parbox{1cm}{\centering In-network\\Fusion}} & \multirow{4}{*}{\parbox{1cm}{\centering{ CogNN\\(proposed)}}} &\multirow{2}{*}{\text{[BIE, FML]}} & 87.04 \\
& & & 87.34 \\
\cmidrule(r){3-4}
 &  & \multirow{2}{*}{\text{[BIE, FI]}} & \bf 88.26\\
& & &  \bf 88.80 \\
\bottomrule
\end{tabular}
\end{center}
\end{table}


\subsubsection{Effectiveness of the Proposed Models}\label{subsub_model}
Since there are no existing work study the fine-grained person name recognition task, we compare with the following deep learning models and variants of our proposed models:

\begin{itemize}
\item \textbf{JointNN}: We share the hidden layers between two tasks while keeping two task-specific output layers, which is similar to \cite{caruana1997multitask}. Specifically, a Bi-LSTM layer is used to get the hidden representations of the input. The hidden representations are passed two task-specific output layers to predict name forms and name span respectively. All the output is further fed into a linear chain CRF layer before perform predictions. And the two tasks are trained jointly to minimize the total loss. Bi-LSTM has a 100-dimensional hidden layer, a dropout layer with probability 0.5, a batch size of 32, and an initial learning rate of 0.01 with a decay rate of 0.05.
\item \textbf{DeepNN}:  We use two successively deeper layers similar to  \cite{sogaard2016deep} for predicting the name form classes and the name spans respectively.  Specifically, two Bi-LSTM layers are stacked for predicting the name form classes and the name spans respectively. The output of the first Bi-LSTM becomes the input of the second Bi-LSTM. All the output is further fed into a linear chain CRF layer to predict the tokens labels. And the two tasks are trained jointly to minimize the total loss. Each Bi-LSTM has a 100-dimensional hidden layer, a dropout layer with probability 0.5, a batch size of 32, and an initial learning rate of 0.01 with a decay rate of 0.05.
\item \textbf{ConcatNN}: We use the concatenating procedure similar to  \cite{ma2018joint} to fuse the output of one task-specific layer with the input of another task-specific layer. Specifically, a Bi-LSTM layer is used to get the hidden representations of the input. The hidden representations are passed to the first task-specific output layer to predict name form classes. Then the output of the name form prediction layer as well as the initial idden representations from Bi-LSTM layer are concatenated as the input of the second task-specific output layer to predict name spans. All the output is further fed into a linear chain CRF layer before perform predictions. And the two tasks are trained jointly to minimize the total loss. Bi-LSTM has a 100-dimensional hidden layer, a dropout layer with probability 0.5, a batch size of 32, and an initial learning rate of 0.01 with a decay rate of 0.05.
\item  \textbf{CoAttNN}: A variant of our proposed CogNN model with co-attention mechanism but not gated-fusion mechanism. The implementation, training procedures and hyperparameters are the same as those for CogNN.
\item \textbf{CogNN}: Our proposed model.
\item \textbf{BERT}: The vanilla version of BERT model. We choose the ``bert-base-case'' as the pre-trained model, which is trained on cased English text and contains 12-layer, 768-hidden-features for each layer and 110M trainable parameters in total. We set the maximum input length as 512.
\item \textbf{IsBERT without multi-hop}:  A variant of our proposed model IsBERT with bidirectional overlapped contextual embedding but not multi-hop inferences. The training procedures and hyperparameters are the same as those for IsBERT.
\item \textbf{IsBERT}: Our proposed model. The training procedures and hyperparameters are the same as those for BERT except that we set the inferences number as 2 and overlapped ratio as 0.5 (256 tokens).
\item \textbf{Ada-IsBERT}: Our proposed model. The training procedures and hyperparameters are the same as those for IsBERT except that we dynamically choose the overlapped ratio from 0, 0.10, 0.20, 0.30, 0.40 and 0.50.
\end{itemize}

All the optimal hyperparameters of above non BERT based deep learning models are obtained with standard grid search on the same developing dataset with stochastic gradient descent. We stop training if the accuracy does not improve in 10 epochs. All the BERT based models are implemented using Pytorch and are trained using Adam optimizer and learning rate decay, starting from $10^{-5}$.

\begin{table}[t!]
\centering
\caption{Token-level and Name-level results of the models evaluated on FinegrainedName}
\begin{tabular}{c|c|c|c|c}
\toprule
\multirow{2}{*}{\textbf{Models}}      & \multicolumn{3}{c|}{\textbf{Token-level}}        & \textbf{Name-level}         \\ \cmidrule(r){2-5}
                                      & \textbf{R}     & \textbf{P}     & \textbf{F}          & \textbf{F}     \\ \midrule
JointNN                               & 88.02          & 89.82          & 88.91                    & 81.04          \\ \midrule
DeepNN                                & 87.84          & 89.44          & 88.63                 & 80.12          \\ \midrule
ConcatNN                               & 87.79          & 89.44          & 88.61           & 80.12          \\ \midrule
CoAttNN                              & 89.23          & 90.54          & 89.88            & 84.26          \\ \midrule
CogNN (proposed)                                & 91.08          & 91.74          & 91.41       & 87.04          \\ \midrule
BERT                                  & 93.17          & 94.55          & 93.85             & 92.76          \\ \midrule
IsBERT no inference & 94.98          & 95.00          & 94.99       & 94.78          \\ \midrule
IsBERT (proposed)   & \textbf{95.27} & \textbf{96.11} & \textbf{95.69}  & \textbf{95.56} \\ \bottomrule
\end{tabular}

\label{result_model}
\end{table}

Table \ref{result_model} shows the results.
JointNN achieves slightly better results at the name level compared with the no fusion Bi-LSTM-CRF model in Table \ref{result_scheme}. However, DeepNN and ConcatNN are worse than no fusion Bi-LSTM-CRF. The reason is that DeepNN and ConcatNN do not have a good mechanism to filter the learning signals. DeepNN uses successively deeper layers to connect two tasks and ConcatNN utilises straightforward concatenating procedures to share the informations. Noisy signals may be introduced into the training process and the propagation of error may reduce the effectiveness of our annotations. Both our proposed model CogNN and its variant CoAttNN outperform these multi-task models. CogNN outperforms the best baseline by up to 5.47\% and 6.06\% in F1-score at the token level and name level, respectively. This verifies the effectiveness of our co-attention and gated fusion mechanisms for utilising the fine-grained annotations. CogNN performs better than CoAttNN since the sub-networks can share the learning signals while neither sub-network is overwhelmed by misleading signals.

All the BERT based model yield better results than GloVe based methods. The vanilla BERT model outperforms the best joint learning model CogNN based on LSTM-CRF by 2.44\% and 5.72\% in F1-score at the token-level and name-level respectively. The transformers inside BERT can better capture the context information compared with the traditional LSTM-CRF based approaches. Our proposed IsBERT model achieves the highest F1-score in both token-level and name-level, which outperforms the CogNN model by 4.28\% and 8.52\% in the field of token and name level respectively. In particular, the IsBERT model achieves a significant improvement in the name-level and maintain roughly consistent performance in both levels. IsBERT also outperforms the naive BERT and IsBERT without multi-hop inferences, which indicates the effectiveness of our proposed bidirectional overlapped contextual embedding learning and multi-hop inferences.

\subsubsection{Influence of Parameters}\label{subsub_para}
\begin{table}[t!]
\centering
\caption{Token-level and Name-level results of IsBERT with different overlapped ratio and different numbers of inference modules on FinegrainedName}
{\begin{tabular}{c|c|c|c|c|c}
\toprule
\multirow{2}{*}{} & \multirow{2}{*}{{\parbox{1cm}{\centering \textbf{Para-\\meters}}}} & \multicolumn{3}{c|}{\textbf{Token-level}} & \textbf{Name-level} \\ \cmidrule{3-6}
                                 &                                            & \textbf{R}  & \textbf{P} & \textbf{F}     &  \textbf{F}     \\ \midrule
\multirow{2}{*}{\parbox{1.5cm}{\centering Overlapped\\Ratio}}   & 0.25                                       & 94.79       & 95.61      & 95.20            & 93.97          \\ \cmidrule{2-6}
                                 & 0.5                                        & 95.27       & 96.11      & \textbf{95.69}   & \textbf{95.56} \\ \midrule
\multirow{3}{*}{\parbox{1.5cm}{\centering Inference\\Modules}}   & 1                     & 94.79       & 95.61      & 95.20             & 93.97          \\ \cmidrule{2-6}
                                                                  &2                                               & 95.27       & 96.11      & \textbf{95.69}     & \textbf{95.56} \\ \cmidrule{2-6}
                                                                  &3                                               & 93.31       & 96.46      & 94.76              & 94.80          \\
                                                                  \bottomrule
\end{tabular}}
\label{tab:or_performance}
\end{table}

\begin{table}[t!]
\centering
\caption{Token-level results of IsBERT and Ada-IsBERT on different abundances of context}
{\begin{tabular}{c|c|c|c|c}
\toprule
 & \textbf{Models} & \textbf{R} & \textbf{P} & \textbf{F}\\
\midrule
\multirow{2}{*}{Short Docs Only} & IsBERT & 79.57	& 85.77 &	82.55 \\
\cmidrule{2-5}
 & Ada-IsBERT & 84.27 &	85.51 &	\textbf{84.88}\\
 \midrule
\multirow{2}{*}{Long Docs Only} & IsBERT & 96.41 &	96.81 &	\textbf{96.61}\\
\cmidrule{2-5}
 & Ada-IsBERT & 96.41 &	96.81 &	\textbf{96.61}\\
 \midrule
\multirow{2}{*}{Mix Length Docs} & IsBERT & 95.04 &	96.09	 &95.56\\
\cmidrule{2-5}
 & Ada-IsBERT & 96.17	& 95.53	& \textbf{95.85} \\
 \bottomrule
 \end{tabular}}
 \label{tab:adaptive}
 \end{table}
%

Table~\ref{tab:or_performance} shows the performance of IsBERT under different overlapped ratios and different numbers of inference modules on long documents. The experiments results meet our expectation for long documents as a higher overlapped ratio leads to better model performance. Compared with an overlapped ratio of 0.25, the F1-score improves about 0.39\% and 1.69\% in token-level and name-level with an overlapped ratio of 0.5. The reason is that more tokens are shared between chunks and better inter-sentence context is captured.

Using two inference modules in IsBERT is optimal for our task, which supports our hypothesis that the multi-hop inference method can strengthen the inherent signals learned from bidirectional overlapped contextual embedding. However, with the number of layers growing, the performance of the model drops and is even lower than the one without using multi-hop inferences. The possible reason is overfitting since the model passes one document for too many times.

Table~\ref{tab:adaptive} shows the performance of IsBERT and Ada-IsBERT on documents with different abundances of context. We select documents with not more than 256 tokens from the FinegrainedName dataset to form the collection of documents with low abundances of context (Short Docs Only). We select documents with more than 256 tokens from the FinegrainedName dataset to form the collection of  documents with high abundances of context (Long Docs Only). We use the original FinegrainedName dataset as the  the collection of  documents with both high and low abundances of context (Mix Length Docs).
The results shows that both IsBERT and Ada-IsBERT experience huge performance drop on Short Docs compared with on Long Docs. The reason is that our models do not have too much context to utilize and to boost the performance. Ada-IsBERT performs much better on Short Docs and Mix Length Docs than IsBERT since it can adjust the overlapped ratio to adapt to documents with different length, and avoid creating repeated context and redundant embeddings in the model.

\subsection{Applicability on Newswire Articles} \label{exp_conll}

While not the focus of this study, we further show the applicability of our models on traditional newswire texts, which are different from academic homepages and mostly have consistent and complete syntax.

We use the CoNLL-2003 dataset which contains 1,393 annotated English newswire articles that focus on four types of named entities: person, location, organisation and miscellaneous entity. We use the training, developing, and testing datasets in CoNLL-2003 to train Stanford NER and IsBERT models with different annotations. The reported improvements are statistically significant with p~$<$~0.05 as calculated using McNemar's test.

This dataset does not come with fine-grained annotations. We add annotations using the same method described in Section \ref{dataset} and compare the following combinations of annotations:
\begin{itemize}
\item \textbf{PER}: Using only PERSON labels.
\item \textbf{FML}: Using only FML labels.
\item \textbf{CoNLL}: Using all original labels in CoNLL-2003.
\item \textbf{FML+CoNLL}: Replacing PERSON by FML labels in CoNLL-2003.
\end{itemize}

\begin{table}[t!]
\begin{center}
\caption{\label{result_CoNLL} Token level performance of person name recognition on CoNLL-2003 using different models.}
\begin{tabular}{c|p{2.95cm}|p{0.5cm}p{0.5cm}p{0.5cm}}
\toprule
\bf{Models} & \centering \bf{Annotations} & \centering{\bf{R}} & \centering{\bf{P}} & \centering{\bf{F}}\tabularnewline
\midrule
\multirow{6}{*}{\parbox{1.5cm}{\centering Stanford NER}} & \centering PER & 85.29 & 94.75 & 89.77\\
& \centering FML & 83.66  & 93.36  & 88.25 \\
& \centering CoNLL & 92.43 & 89.96 & 91.18\\
& \centering FML+CoNLL & 90.19 &  89.04 & 89.61 \\
\midrule
\multirow{6}{*}{\parbox{1.5cm}{\centering IsBERT \\(proposed)}}& \centering PER & 94.78 & 92.82& 93.79\\
& \centering FML & 95.27 & 96.11 &  95.69\\
& \centering CoNLL & 94.78 & 92.82  &  93.79\\
& \centering FML+CoNLL & 92.34 & 94.11 & 93.23 \\
\bottomrule
\end{tabular}
\end{center}

\end{table}

From Table \ref{result_CoNLL},
we see that neural models perform better than the non-neural model, which is consistent with the results in Section \ref{subsub_scheme}. When providing extra ORG, LOC, and MISC annotations apart from PER to Stanford NER, the recall increases while the precision decreases. This indicates that the extra annotations help recognise more named entity tokens but may also misguide the model. In comparison, IsBERT is not impacted.
When providing extra FML annotations apart from PER to Stanford NER the performance does not improve while that of IsBERT improves. Our improvements mainly lie in the precision, with a 3.29\% improvement at the token level compared with the best baseline, which indicates that IsBERT can well distinguish person name tokens from others. These results also indicate that only applying the fine-grained name form annotations on newswire data for the existing models is not enough. Our proposed model is essential to make use of the extra name form information.

Overall, our approach is also applicable on formal English newswire articles and is especially helpful for improving the precision. However, the advantage of our approach is smaller than that on the academic homepages. The main reason is that the name forms in newswire articles are less flexible compared with those in academic homepages, which reduces the benefits of adding extra name form information.

\begin{figure}[t!]
  \centering
  \includegraphics[width=0.7\columnwidth]{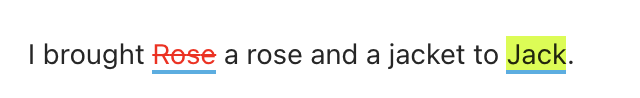}
  \caption{Stanford NER on short text}
  \label{fig:case1}
\end{figure}

\begin{figure}[t!]
  \centering
  \includegraphics[width=0.65\columnwidth]{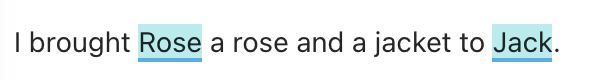}
  \caption{Ada-IsBERT on short text}
  \label{fig:case2}
\end{figure}

\begin{figure}[t!]
  \centering
  \includegraphics[width=\columnwidth]{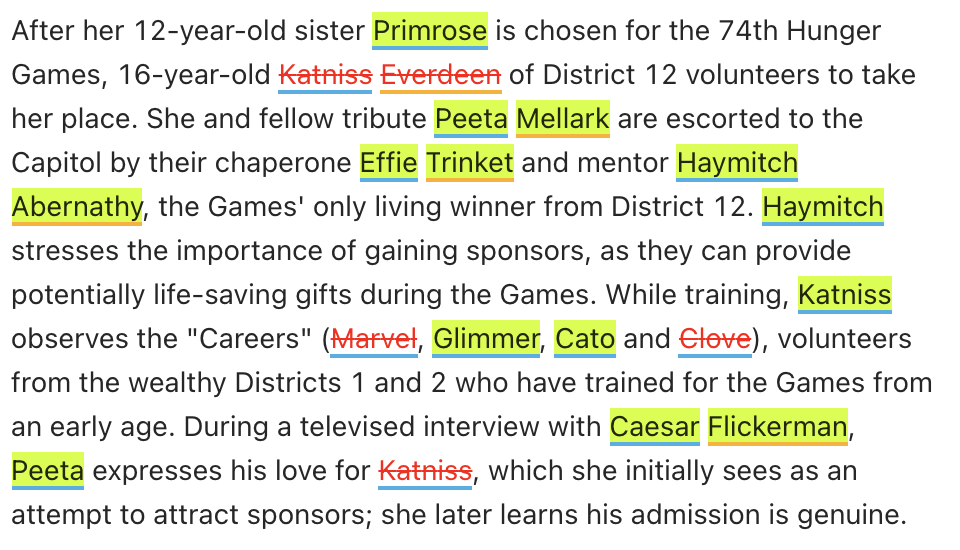}
  \caption{Stanford NER on well-formatted text}
  \label{fig:case3}
\end{figure}

\begin{figure}[t!]
  \centering
  \includegraphics[width=\columnwidth]{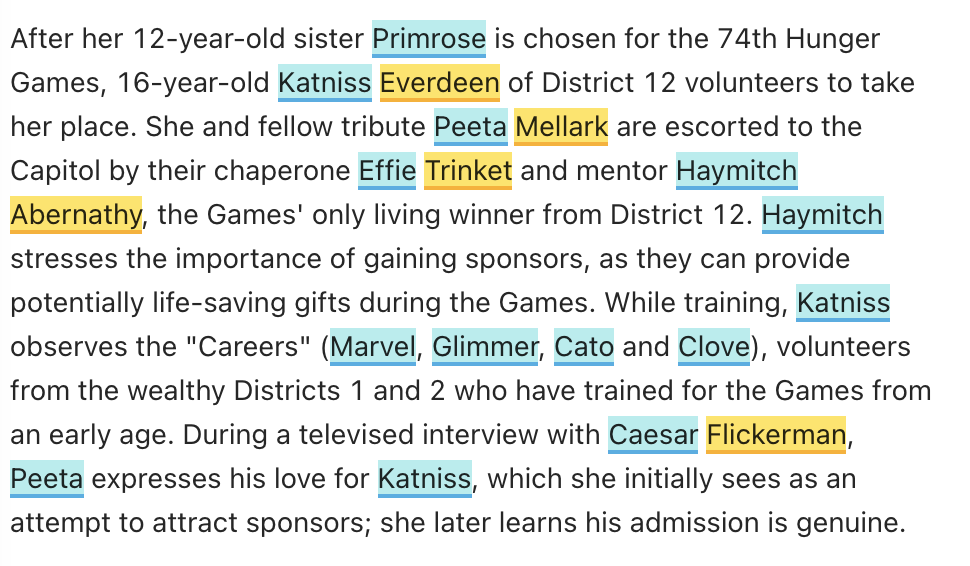}
  \caption{Ada-IsBERT on well-formatted text}
  \label{fig:case4}
\end{figure}

Figure~\ref{fig:case1},~\ref{fig:case2},~\ref{fig:case3},~\ref{fig:case4} show the results of Stanford NER and Ada-IsBERT on different cases. Model predictions are highlighted using the background colour, while the underline colour shows the ground truths. False negative and false positive predictions are noted by red crossline.
We can see from the Figure~~\ref{fig:case3},~\ref{fig:case4} that our proposed model can not only recognize more person names than Stanford NER, our model also provide fine-grained name forms with higher accuracy. We also test two models on short text, Figure~\ref{fig:case1},~\ref{fig:case2} show that our model also perfroms well on text without too much context, and it can distinguish the same tokens with different meanings.

\subsection{Error Analysis} \label{exp_case}

Some typical errors of the proposed model are investigated and analyzed in this section. Model predictions are highlighted using the background colour, while the underline colour shows the ground truths. False negative and false positive predictions are noted by red crossline.

\begin{figure}[t!]
  \centering
  \includegraphics[width=\columnwidth]{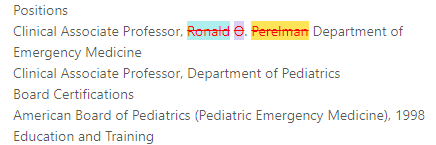}
  \caption{Errors in organization names}
  \label{fig:nr_error4}
\end{figure}

 \textbf{Errors in organization names}: Figure~\ref{fig:nr_error4} shows a typical error that the model labels the name of an organization as a person name. Even though the organization is named by a person ``Ronald O. Perelman'', we expecte that the model could distinguish organization names from person names by the context information, such as the ``Department of Emergency Medicine''. It is  challenging for the model to perform well in this situation. To solve this problem, we could enhence our trainng dataset by addding more training examples that containing organizations named by person names labelled as negative. We leave this for future works.

\begin{figure}[t!]
  \centering
  \includegraphics[width=\columnwidth]{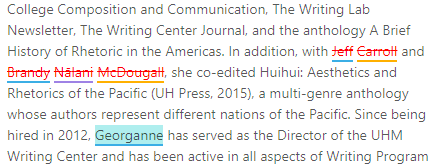}
  \caption{Errors in well-formatted text}
  \label{fig:nr_error1}
\end{figure}

\begin{figure}[t!]
  \centering
  \includegraphics[width=\columnwidth]{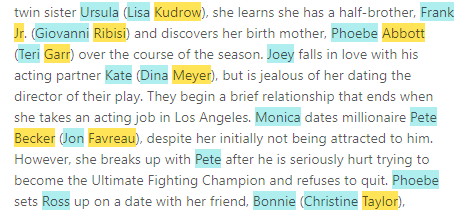}
  \caption{Example with enhancement data}
  \label{fig:nr_error5}
\end{figure}

\textbf{Errors in well-formatted text}: Figure~\ref{fig:nr_error1} shows a typical error that the model does not recognize some person names in the well-formatted text, such as in the introductions or in the career descriptions. The reason is that, in the FinegrainedName dataset, most person names appear in the publications and come with different formats as those the introductions. Without sufficient training data that containing person names in the well-formatted text, the model tends to miss these person names. To solve this problem, we could enhence our trainng dataset by adding more well-formatted text articles such as CONLL-2003 dataset. We leave this for future works.
To solve this problem, we add 200 annotated news articles from the CONLL-2003 dataset to the training set to balance the training data.
Figure~\ref{fig:nr_error5} shows an example which is tested on the well-formatted text after the data enhancement. IsBERT recognize all the person names correctly.


%

\section{Conclusion} \label{conclusion}
We studied the person name recognition problem. We propose a new name annotation scheme which gives fine-grained name annotations and a new model called CogNN to take advantage of the fine-grained annotation scheme via co-attention and gated fusion. We have also created the first dataset under the fine-grained annotation scheme, called FinegrainedName, for the research of name recognition. We also propose a IsBERT to recognise fine-grained person names in long documents by capturing the inter-sentence context and implicit relations via an overlapped input processor, and an inter-sentence encoder with bidirectional overlapped contextual embedding learning and multi-hop inference mechanisms. We also propose an Ada-IsBERT which is robust for documents with diverse abundances of context by dynamically adjusting and  the inter-sentence overlapping ratio according to the length of different documents.

Experiments on FinegrainedName dataset and Newswire dataset show that our annotations can be utilised in different ways to improve the person name recognition performance in many application cases with different document length. Our CogNN model outperforms other NER models and multi-tasks models on utilising the fine-grained name form information. Our IsBERT model outperforms other models on capturing the inter-sentence context in long documents. Our Ada-IsBERT model is robuster for documents with different abundances of context. All our models outperforms state-of-the-art NER models for person name recognition and is especially advantageous in having high precision.

For future work, we plan to investigate the fine-grained annotations and the proposed models on other languages rather than English.
We also plan to evaluate the performance of our approach on other types of datasets such as online forums and social media.

%
%
%
%

\bibliographystyle{IEEEtran}
\bibliography{IEEEabrv,IEEEexample}

\end{document}